# A newborn embodied Turing test for view-invariant object recognition

Denizhan Pak[1-2], Donsuk Lee[1-2], Samantha M. W. Wood[1,3], Justin N. Wood[1-4]
(denpak@iu.edu; donslee@iu.edu; sw113@iu.edu; woodjn@iu.edu)
[1]Department of Informatics,
[2]Cognitive Science Program,
[3]Department of Neuroscience,
[4]Center for the Integrated Study of Animal Behavior,
Indiana University, Bloomington, IN 47408 USA

**Abstract**

Recent progress in artificial intelligence has renewed interest in building machines that learn like animals. Almost all of the work comparing learning across biological and artificial systems comes from studies where animals and machines received different training data, obscuring whether differences between animals and machines emerged from differences in learning mechanisms versus training data. We present an experimental approach—a "newborn embodied Turing Test"—that allows newborn animals and machines to be raised in the same environments and tested with the same tasks, permitting direct comparison of their learning abilities. To make this platform, we first collected controlled-rearing data from newborn chicks, then performed "digital twin" experiments in which machines were raised in virtual environments that mimicked the rearing conditions of the chicks. We found that (1) machines (deep reinforcement learning agents with intrinsic motivation) can spontaneously develop visually guided preference behavior, akin to imprinting in newborn chicks, and (2) machines are still far from newborn-level performance on object recognition tasks. Almost all of the chicks developed view-invariant object recognition, whereas the machines tended to develop view-dependent recognition. The learning outcomes were also far more constrained in the chicks versus machines. Ultimately, we anticipate that this approach will help researchers develop embodied AI systems that learn like newborn animals.

## Introduction

Since the birth of artificial intelligence (AI), scientists have attempted to build machines that can learn like biological systems. Early AI research laid the foundation for biologically inspired, neurally mechanistic models, and recent progress in deep learning has renewed interest in building scalable AI systems that learn like animals. For instance, a new "reverse engineering" paradigm in computational neuroscience involves comparing neural and behavioral measurements across animals and artificial systems performing the same tasks. Reverse engineering has revolutionized our algorithmic understanding of vision (Yamins & DiCarlo, 2016), audition (Kell et al., 2018), olfaction (Wang et al., 2021), and visually-guided action (Michaels et al., 2020), while also informing our understanding of higher-level cognitive abilities—including language (Schrimpf et al., 2021), navigation (Whittington et al., 2022), and memory (Nayebi et al., 2021).

Ultimately, how will we know when we have succeeded in building machines that learn like animals? To address this question, we argue that an experimental platform must have two core features: (1) the animals and machines must be raised in the same environments; (2) the animals and machines must be tested with the same tasks. The first requirement follows from the observation that behavior depends both on the *learning mechanisms* and the *training data* on which the mechanisms operate. Any observed differences in behavior across animals and machines could be due to differences in learning mechanisms, training data, or some combination of the two. Thus, evaluating whether machines learn like animals requires giving machines the same training data (experiences) as animals. The second requirement follows from the observation that evaluations of intelligence and learning are task-dependent. Accordingly, biological and artificial systems must be evaluated with the same tasks to ensure that any observed differences are not due to differences in the tasks themselves.

While these two requirements may seem straightforward, building an experimental platform that meets both requirements has not previously been possible. Controlling the training data across animals and machines requires performing parallel controlled-rearing experiments on animals and machines. However, most newborn animals cannot be raised in controlled environments from birth, preventing researchers from controlling the training data presented to animals. Accurate comparison between animals and machines also requires measurements with a high signal-to-noise ratio, where a subject's behavior in response to a particular stimulus (e.g., an image) can be reliably estimated. However, most prior controlled-rearing studies collected data with a low signal-to-noise ratio and focused on group-level analyses across coarse experimental conditions. As a result, the field lacked the high-precision data needed to make accurate comparisons across newborn animals and machines. Finally, the field lacked an experimental platform for raising machines in the same environments as newborn animals, preventing researchers from matching the training data across biological and artificial systems.

Here we present an experimental approach that overcomes these barriers, allowing newborn animals and machines to be raised in the same environments and tested with the same tasks (Figure 1). We used newborn chicks as a model system because they are mobile on the first day of life and can be raised in strictly controlled environments from the onset of vision (Wood & Wood, 2015). As a starting point, we focused on building machines that can mimic the *imprinting behavior* of newborn chicks. We focused on imprinting because (a) it is one of the earliest forms of visual learning that can be studied with high precision in a biological system (Wood & Wood, 2015), (b) it produces powerful (invariant) representations that support object recognition across new viewing situations (Wood, 2013; Wood & Wood, 2021), and (c) it emerges spontaneously during an animal's early interactions with the world, driven by self-organized learning mechanisms. There is growing demand in AI for self-organizing systems that can learn from sparse data. Imprinting is therefore a promising target for reverse engineering the development of visual intelligence in embodied systems.

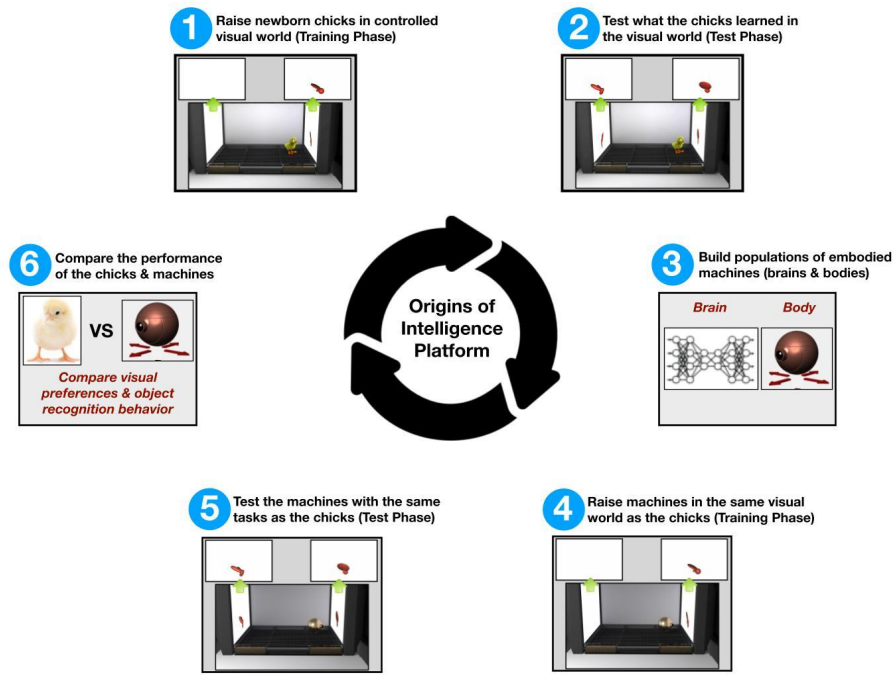

**Figure 1.** Experimental approach for comparing the learning abilities of newborn chicks and machines. The animals and machines are raised in the same visual environments and tested with the same tasks. This "newborn embodied Turing test" evaluates whether animals and machines develop the same visual behaviors when provided with the same training data.

In this paper, we tested whether machines can mimic the imprinting and object recognition behavior of newborn chicks. Specifically, we focused on the controlled-rearing experiments from Wood (2013). Wood raised newborn chicks in controlled visual environments containing a single virtual object, then measured the chicks' imprinting response and view-invariant recognition performance. To obtain data with a high signal-to-noise ratio, Wood (2013) used an automated controlled-rearing method that measured the chicks' behavior continuously (24/7). In the current study, we created digital twins (virtual environments) of the animal chambers in a video game engine and raised autonomous machines in those virtual chambers. By raising animals and machines in the same visual environments, we could measure whether they spontaneously develop common visually guided behaviors. We compared newborn chicks and machines on two measures:

- **Object Preference Behavior.** To mimic the chicks, the machines should develop a preference for the imprinted object, without any explicit rewards or supervision. The machines must also develop knowledge of their location and direction in space (ego-motion), so that they can navigate to their imprinted (preferred) object.
- **Object Recognition Behavior.** To mimic the chicks, the machines should learn to recognize the imprinted object across novel views (Figure 2). This requires learning view-invariant object features in impoverished visual environments containing a single object seen from a limited range of views, using a purely self-supervised learning strategy (i.e., no supervised labels or rewards).

### Animal Experiments

In the study, chicks were hatched in darkness, then raised singly in automated controlled-rearing chambers for the first two weeks of life. The chambers contained two display walls (LCD monitors) for displaying object stimuli (Figure 1). The chambers did not contain any objects other than the virtual objects projected on the display walls. Thus, the chambers provided full control over all visual object experiences available to the chicks from the onset of vision.

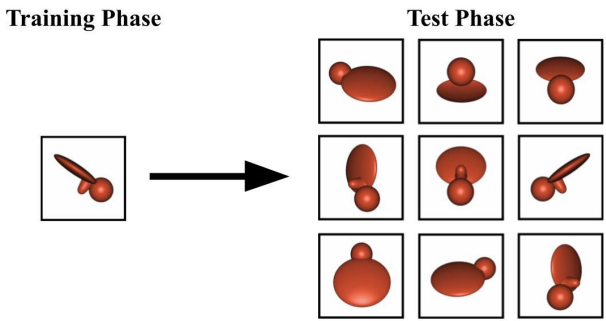

**Figure 2.** View-invariant recognition task. Newborn chicks and machines were reared in environments containing a single object seen from a single viewpoint range, then tested on their ability to recognize that object across novel views.

During the Training Phase, the chicks were reared with a single 3-D object rotating through a 60° viewpoint range. The object rocked back and forth every 6s. The chicks were reared with one of two possible objects, and the object was presented from a front or side viewpoint range (Figure 3A). The chicks were raised in this environment for one week, allowing the critical period on filial imprinting to close.

During the Test Phase, the chambers measured the chicks' imprinting response and object recognition performance. The "imprinting trials" measured whether the chicks developed an imprinting response. During these trials, the imprinting stimulus was presented on one display wall and the other display wall was blank. If the chicks developed an imprinting response, then they should have spent more time by the display wall showing the imprinting stimulus than the blank display wall. The "test trials" measured the chicks' view-invariant recognition performance. During these trials, the imprinted object was presented on one display wall and an unfamiliar object was presented on the other display wall. Across trials, the imprinted object was presented from 12 viewpoint ranges (11 novel, 1 familiar). Conversely, the unfamiliar object (which was the same size and color as the imprinted object) was presented from the same viewpoint range as the imprinted object from the input phase (for details see Wood, 2013). Consequently, on most of the test trials, the unfamiliar object was more similar to the imprinting stimulus than the imprinted object was to the imprinting stimulus (from a pixel-wise perspective). To recognize their imprinted object, the chicks therefore needed to generalize across large, novel, and complex changes in the object's appearance. This meets a reasonable operational definition of "invariant object recognition" established with mature animals (Zoccolan et al., 2009). If the chicks developed view-invariant recognition, then they should have spent more time by the display wall showing the imprinted object than the display wall showing the unfamiliar object.

| Table 1: Hyperparameters | |
|---|---|
| **Policy Network (PPO)** | **Intrinsic Motivation Networks** |
| vis_encode_type: 2 convolutional layers | vis_encode_type: 2 convolutional layers |
| num_layers: 2 | num_layers: 2 |
| hidden_units: 128 | hidden_units: 128 |
| learning_rate: 0.0003 | learning_rate: 0.0003 |
| batch_size: 500 | strength: 1.0 |
| buffer_size: 2048 | gamma: 0.99 |
| beta: 0.01 | |
| epsilon: 0.2 | |
| lambda: 0.95 | |
| learning_rate_schedule: linear | |
| max_steps: 1000000 | |

## Machine Experiments

Our primary goal was to raise and test animals and machines under parallel conditions. This required (1) machines that can learn from raw sensory inputs and perform actions, akin to real animals, and (2) virtual environments for machines that mimic the visual environments of the chicks.

**Pixels-to-Actions Machines.** Newborn animals learn from raw sensory inputs and perform actions, driven by self-supervised learning objectives. Thus, to directly compare animals and machines, we used 'pixels-to-actions' machines that learn from raw sensory inputs and perform actions, driven by self-supervised learning objectives (e.g., intrinsic motivation).

We created the machines by embodying self-supervised learning algorithms (Figure 3D) in virtual bodies. Each body measured 3.5 units (height) by 1.2 units length (radius) and received visual input (96×96 pixel resolution images) through an invisible forward-facing camera attached to its head. The machines could move forward or backwards, rotate left or right, or remain stationary. The actions were represented as a pair of discrete variables: translation forward/backward and rotation around the vertical axis.

As a starting point, we built the machine brains using a simple 2-layer convolutional visual encoder and a standard reinforcement learning system: Proximal Policy Optimization (PPO; Schulman et al., 2017). During the Training Phase, the PPO algorithm was optimized for rewards generated by one of three intrinsic motivation algorithms: Intrinsic Curiosity (Pathak et al., 2017), Random Network Distillation (Burda et al., 2018), or Contrastive Curiosity (Nguyen et al., 2021) adapted from the SimCLR learning algorithm (Chen et al., 2020). Each algorithm takes batches of inputs and produces rewards. The batch and reward are then used to train the PPO network.

All three of the intrinsic motivation algorithms produced rewards based on the "novelty" of the input, with more unique inputs generating greater rewards. The Intrinsic Curiosity algorithm generated rewards based on the machine's ability to predict the next state given the current state and action. The Random Network Distillation algorithm generated rewards based on whether the machine could predict the embedding generated by an input in a random network. And the Contrastive Curiosity algorithm generated rewards based on the distance (in the embedding space) between current inputs and prior inputs, using a contrastive learning scheme. We created our agents using Unity ML-Agents Toolkit v. 2.10.1 (Juliani et al., 2018).

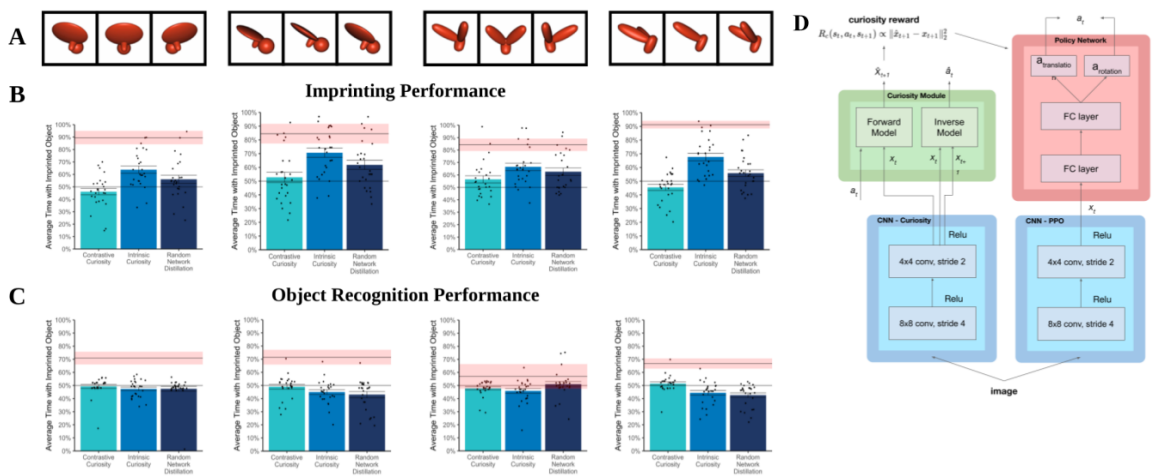

**Figure 3. (A)** The imprinting stimuli presented to the chicks and machines in the four rearing conditions. The object rocked back and forth through a 60° viewpoint range. **(B)** Imprinting performance of the chicks and machines. Mean chick performance is shown as a red line with a "noise band" (red band) reflecting the average deviation of the chicks from the mean chick performance. Mean machine performance is shown in the bars with standard error across machines. Mean performance of individual machines is shown as dots (the mean standard error for individual machines was 7% for imprinting and 3% for object recognition). If the machines developed similar learning outcomes as the chicks, then the mean machine behavior should fall within the noise band. There was a large gap between the imprinting performance of the chicks and machines. **(C)** Object recognition performance of the chicks and machines. For three of the four rearing conditions, there was a large gap between the object recognition performance of the chicks and machines. **(D)** Architecture of the Intrinsic Curiosity PPO learning algorithm. The blue boxes denote visual encoders (CNNs), the green box denotes the curiosity module (intrinsic reward), and the red box denotes the policy network used to select actions.

**Virtual Environments for Machines.** To simulate the visual environments of the chicks in Wood (2013), we raised (trained) the machines in realistic digital twins of the controlled-rearing chambers, using a video game engine (Unity 3D; Figure 1). We then tested the machines in the virtual chambers, presenting the same stimuli and tasks to the machines that were presented to the chicks. For each of the four rearing conditions in Wood (2013), we trained and tested 26 machine subjects for each intrinsic motivation algorithm. All machines had the same network architecture: 2-layer CNN connected to a multilayer perceptron (see Table 1 for hyperparameters). However, each machine's neural network started with a different random initialization of connection weights, and each machine's connection weights were updated based on its own particular experiences during the Training Phase. Like chicks, the machines received no external rewards from the environment. The actions were motivated entirely by the rewards from their intrinsic motivation algorithm.

At the beginning of each training episode, the machines were spawned at a random position and orientation within the chamber. The training episodes lasted 1,000 time steps. We trained the machines for 1,000 episodes. The machines were trained to optimize the sum of their intrinsic motivation reward using PPO.

After the Training Phase, the network weights were frozen for the Test Phase (i.e., the machines did not receive any rewards during the Test Phase, and learning was discontinued). This freezing mimics the critical period of filial imprinting in chicks, in which chicks stop learning about their imprinted object after the first few days of life. Each machine performed 480 test trials (40 trials for each of the 12 viewpoint ranges presented to the chicks). Each test trial consisted of 1,000 time steps. At every time step, we recorded the position of the machine in X,Y coordinates. As with the chicks, we measured whether the machines spent a greater proportion of time with the imprinted object than the unfamiliar object. Since we only considered position and not head orientation in our measure, we also confirmed that the agents tended to look where they moved by analyzing the difference between their head orientation and movement direction. All three of the algorithms tended to look where they moved.

## Results

**Imprinting results.** Figure 3B shows chick and machine performance on the imprinting trials. On the group level, the chicks spent significantly more time by the imprinting stimulus than the blank screen ($M = 88\%$, SD = 7%, $t(22) = 25.74$, $p < .0001$). On the individual level, all of the chicks successfully learned to imprint (range 72% to 97%, chance performance = 50%).

Likewise, on the group level, the machines spent significantly more time by the imprinting stimulus than the blank screen (all machines: $M = 59\%$, SD = 17%, $t(311) = 9.22$, $p < .0001$; Contrastive Curiosity: $M = 50\%$, SD = 15%, $t(103) = 0.13$, $p = .90$; Intrinsic Curiosity: $M = 67\%$, SD = 15%, $t(103) = 11.41$, $p < .0001$; Random Network Distillation: $M = 59\%$, SD = 16%, $t(103) = 5.93$, $p < .0001$). Some of the machines developed an imprinting response, spontaneously learning to seek out the imprinting stimulus. However, on the individual level, imprinting was far less robust in the machines compared to the chicks. While 100% of the chicks learned to imprint, only a minority (37%) of the machines learned to imprint. Some intrinsic motivation algorithms produced imprinting behavior more often than others, but no algorithm produced consistent imprinting. In all cases, the machines developed markedly different imprinting responses from one another, despite the machines starting with the same learning mechanisms and learning in the same visual environment. In contrast, the newborn chicks universally developed robust imprinting behavior, indicating that there is more variation in the development of machines versus animals.

**Object recognition results.** Figure 3C shows overall chick and machine performance on the test trials. The large majority of chicks (87%) developed view-invariant object recognition ($P$s < .05), successfully recognizing their imprinted object across novel views. Thus, newborn chicks can learn to recognize objects across large, novel, and complex changes in the object's appearance (Wood, 2013). A small handful of the machines (5%) also developed view-invariant recognition, demonstrating that it is possible for machines to learn the same task as newborn chicks. However, it was more common (21%) for the machines to develop view-dependent recognition, favoring the novel object presented on the test trials (which was a closer match to the imprinting stimulus from a pixel-level perspective). To visualize the recognition differences between the chicks and machines across the 12 viewpoint ranges, we plotted a t-SNE embedding (Hinton & Roweis, 2002). The t-SNE algorithm revealed substantial differences in learning outcomes across chicks and machines (Figure 4). The chicks (red dots) were largely clustered in the same part of the space, indicating that chicks developed similar recognition behavior as one another. Conversely, all three of the machine-learning algorithms (blue dots) generated highly variable recognition performance, highlighting the considerable gap in object recognition behavior across the animals and machines.

**Animal-Machine Performance Gap.** To quantify the gap in performance between the chicks and machines, we computed a "noise band" around the mean performance of the chicks (Figures 3B & 3C). The noise band reflects the average deviation of the chicks from mean chick performance. This defines a clear noise ceiling to judge prediction accuracy (Cao & Yamins, 2021). Noise ceilings capture the idea that the accuracy of a machine in predicting chick behavior can only be as good as the accuracy of a chick in predicting chick behavior (e.g., due to measurement error and individual differences across chicks). If average machine performance reaches the noise band, then the machines can be said to be "predictively adequate" of chick development (i.e., the learning algorithms in machines generate similar learning outcomes as the learning algorithms in chicks). This was not the case: there were large gaps in imprinting and object recognition performance across the chicks and machines.

## Discussion

We performed digital twin experiments, in which newborn chicks and machines were raised and tested in the same visual environments. This approach permits direct comparison of whether animals and machines learn the same behaviors when provided with the same experiences (training data). In this paper, we explored whether self-supervised machines can spontaneously learn visual preferences (i.e., imprinting) and view-invariant object recognition, mimicking the early emerging behaviors of newborn chicks.

We found that imprinting can emerge in deep reinforcement learning machines equipped with intrinsic motivation. However, unlike newborn chicks, only a subset of the machines successfully imprinted. Object recognition performance also differed significantly across the chicks and machines. A large majority of the chicks developed view-invariant recognition, successfully recognizing their imprinted object across large, novel, and complex changes in the object's appearance. In contrast, many of the machines developed view-dependent recognition, preferring the visual stimulus that was the closest match (from a pixel-level perspective) to the imprinting stimulus. These results indicate that while deep reinforcement learning and intrinsic motivation are sufficient for developing rudimentary forms of imprinting and object recognition, a large gap still exists between the visual learning abilities of newborn chicks and machines.

How might we close this gap? Prior work shows that artificial visual systems (self-supervised CNNs like the visual encoders we used here) can successfully learn view-invariant object features in these visual environments (Lee, Pak, & Wood, 2021; Lee, Gujarathi, & Wood, 2021). This finding suggests that the CNN 'front-end' (visual system) of our agents is not the cause of this performance gap. We suspect that closing the gap between animals and machines will require innovations on the 'back end' of the algorithm (e.g., dynamical processes related to memory, decision making, action, and agency). Our experimental approach should be useful for exploring these possibilities, enabling researchers to systematically test whether AI algorithms can learn visually-guided behaviors as rapidly and efficiently as newborn chicks. For example, researchers might explore how different architectures, objective functions, and learning rules perform on this task (Richards et al., 2019). Since our approach focuses on embodied agents, researchers might also explore how changes in embodiment (e.g., action space, morphology, sensor types) and learning dynamics (e.g., type of intrinsic motivation, metabolic pressures, network dynamics, lifelong learning) influence the behaviors learned by machines.

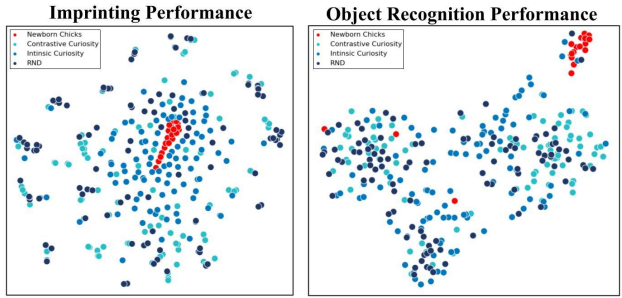

**Figure 4.** Two-dimensional t-SNE embedding of the object recognition behavior across the chicks (red dots) and machines (blue dots). Recognition behavior was measured in a 12-dimensional space, with each dimension reflecting the agent's performance on one of the 12 viewpoint ranges. The machines developed a wide variety of object recognition behaviors, despite having identical neural architectures. Conversely, the chicks developed common object recognition behavior, with most of the chicks developing view-invariant object recognition. These results indicate that the learning mechanisms in newborn chicks, but not those in the machines, constrain visual learning to adaptive parts of the parameter space.

These results also suggest that reverse engineering newborn intelligence will require comparing the *variation of learning outcomes* produced by biological versus artificial systems. The biological algorithms produced view-invariant learning outcomes in most (87%) of the chicks, whereas the artificial algorithms produced view-invariant learning outcomes in a small fraction (5%) of the machines. Thus, the learning outcomes were far more constrained in the chicks versus machines (Figure 4). We emphasize that the machines developed a wide variety of behaviors despite (1) having identical neural architectures and (2) being raised in the same simple visual environments (i.e., environments with four white walls and a single virtual object). We speculate that a core signature of biological intelligence is that it contains developmental programs that produce constrained learning outcomes in populations of agents.

**Reverse engineering embodied learning systems.** Our general contribution is to introduce an experimental approach for reverse engineering visual intelligence in a newborn model system. This approach has much in common with the reverse-engineering approach that revolutionized the neuroscience of perception, including a reliance on precise (high signal-to-noise ratio) data from biological systems and a shared goal of building neurally mechanistic, image computable models of visual intelligence (Schrimpf et al., 2020). While we did not focus on internal (neural) measurements here, future research could expand this digital twin approach to include neural measurements from newborn animals.

Our approach also prioritizes different dimensions of the reverse-engineering problem. We focus on newborn animals (rather than mature animals) in order to study the core learning mechanisms that power visual intelligence. We focus on controlled rearing (rather than animals raised in natural worlds) in order to understand how core learning mechanisms and visual experience interact to produce visual intelligence. And we focus on embodied (rather than disembodied) AI systems, embracing the possibility that much of visual intelligence might emerge from an agent's interactions with the world, allowing learning systems to ground knowledge in real-world experiences and interact with the environment in purposeful ways. Our approach thus extends the call from a large group of scientists arguing for "embodied Turing tests" that involve benchmarking and comparing how animals versus machines learn and interact with the world (Zador et al., 2023).

## Conclusion

We have shown how advances from diverse fields can be linked for comparing newborn animals and machines side-by-side in the same learning settings: (1) automated controlled rearing allows precise data to be collected from newborn animals; (2) video game engines allow machines to be raised in realistic visual environments; (3) AI provides scalable and embodied (pixels-to-actions) learning systems; and (4) computational neuroscience offers a reverse engineering framework for interpreting parallel studies of biological and artificial systems. By combining advances across fields, we can explore which learning mechanisms are necessary and sufficient to mimic the powerful and flexible learning capacities of newborn animals.

Ultimately, we anticipate that a machine with the same learning mechanisms (brain and body) and training data (environment) as newborn chicks should pass this newborn embodied Turing test, developing the same visual preferences and object recognition abilities as newborn chicks. Our hope is that this approach will empower researchers to build machines (and engineering-level scientific models) that learn like newborn animals.

## Acknowledgments
Funded by NSF CAREER Grant BCS-1351892 and a James S. McDonnell Foundation Understanding Human Cognition Scholar Award.